# Underwater target detection based on improved YOLOv7

**Kaiyue Liu [1], Qi Sun [2], Daming Sun [2], Mengduo Yang [3,4,*] and Nizhuan Wang [1,5,*]**

[1] School of Marine Technology and Geomatics, Jiangsu Ocean University, Lianyungang 222023, China
[2] Beijing KnowYou Technology Co., Ltd., Beijing 100086, China
[3] School of Computer Engineering, Jiangsu Ocean University, Lianyungang 222023, China
[4] School of Information Technology, Suzhou Institute of Trade & Commerce, Suzhou 215009, China
[5] School of Biomedical Engineering, ShanghaiTech University, Shanghai 201210, China
* Correspondence: wangnizhuan1120@gmail.com (NW) or mengduoyang@163.com (MY);

**Abstract:** Underwater target detection is a crucial aspect of ocean exploration. However, conventional underwater target detection methods face several challenges such as inaccurate feature extraction, slow detection speed and lack of robustness in complex underwater environments. To address these limitations, this study proposes an improved YOLOv7 network (YOLOv7-AC) for underwater target detection. The proposed network utilizes an ACmixBlock module to replace the 3x3 convolution block in the E-ELAN structure, and incorporates jump connections and 1x1 convolution architecture between ACmixBlock modules to improve feature extraction and network reasoning speed. Additionally, a ResNet-ACmix module is designed to avoid feature information loss and reduce computation, while a Global Attention Mechanism (GAM) is inserted in the backbone and head parts of the model to improve feature extraction. Furthermore, the K-means++ algorithm is used instead of K-means to obtain anchor boxes and enhance model accuracy. Experimental results show that the improved YOLOv7 network outperforms the original YOLOv7 model and other popular underwater target detection methods. The proposed network achieved a mean average precision (mAP) value of 89.6% and 97.4% on the URPC dataset and Brackish dataset, respectively, and demonstrated a higher frame per second (FPS) compared to the original YOLOv7 model. The source code for this study is publicly available at https://github.com/NZWANG/YOLOV7-AC. In conclusion, the improved YOLOv7 network proposed in this study represents a promising solution for underwater target detection and holds great potential for practical applications in various underwater tasks.

**Keywords:** Underwater target detection; Marine resources; Computer vision; Image analysis; YOLOv7-AC; GAM; K-means++

## 1. Introduction

The oceans occupy a significant portion of the Earth's surface and are a valuable source of oil, gas, minerals, chemicals, and other aquatic resources, attracting the attention of professionals, adventurers, and researchers, leading to an increase in marine exploration activities [1]. To support these exploration efforts, various underwater tasks such as target location, biometric identification, archaeology, object search, environmental monitoring, and equipment maintenance must be performed [2]. In this context, underwater target detection technology plays a crucial role. Underwater target detection can be categorized into acoustic system detection and optical system detection [3], and image analysis, including classification, identification, and detection, can be performed based on the obtained image information. Optical images, compared to acoustic images, offer higher resolution and a greater volume of information and are more cost-effective in terms of acquisition methods [4, 5]. As a result, underwater target detection based on optical systems is receiving increased attention.Target detection, being as a core branch of computer vision, encompasses fundamental tasks such as target classification and localization. The existing approaches to target detection can be broadly classified into two categories: traditional target detection methods and deep learning-based target detection methods [6].

Traditional algorithms for target detection are typically structured into three phases: region selection, feature extraction, and feature classification [7]. The goal of region selection is to localize the target, as the position and aspect ratio of the target may vary in the image. This phase is typically performed by traversing the entire image using a sliding window strategy [8], wherein different scales and aspect ratios are considered. Subsequently, feature extraction algorithms such as Histogram of Oriented Gradients (HOG) [9] and Scale Invariant Feature Transform (SIFT) [10] are employed to extract relevant features. Finally, the extracted features are classified using classifiers such as Support Vector Machines (SVM) [11] and Adaptive Boosting (Ada-Boost) [12]. However, the traditional target detection method has two major limitations: (1) the region selection using sliding windows lacks specificity and leads to high time complexity and redundant windows, and (2) the hand-designed features are not robust to variations in pose.

The advent of deep learning has revolutionized the field of target detection and has been extensively applied in computer vision. Convolutional neural networks (CNNs) have demonstrated their superior ability to extract and model features for target detection tasks, and numerous studies have demonstrated that deep learning-based methods outperform traditional methods relying on hand-designed features [13]. Currently, there are two main categories of deep learning-based target detection algorithms: region proposal-based algorithms and regression-based algorithms. The former category, also referred to as Two-Stage target detection methods, are based on the principle of coarse localization and fine classification, where candidate regions containing targets are first identified and then classified. Representative algorithms in this category include R-FCN (Region-based Fully Convolutional Networks) [15] and the R-CNN (Region-CNN) family of algorithms (R-CNN [16], Fast-RCNN [17], Faster-RCNN [18], Mask-RCNN [19], Cascade-RCNN [20], etc.). Although region-based algorithms have high accuracy, they tend to be slower and may not be suitable for real-time applications. In contrast, regression-based target detection algorithms, also known as One-Stage target detection algorithms, directly extract features through CNNs for the prediction of target classification and localization. Representative algorithms in this category include the SSD (Single Shot MultiBox Detector) [21] and the YOLO (You Only Look Once) family of algorithms (YOLO [23], YOLO9000 [24], YOLOv3 [25], YOLOv4 [26], YOLOv5 [27], YOLOv6 [28], YOLOv7 [29]). Due to the direct prediction of classification and localization, these algorithms offer a faster detection speed, making them a popular research area in the field of target detection, with ongoing efforts aimed at improving their accuracy and performance.

The commercial viability of underwater robots equipped with highly efficient and accurate target detection algorithms is being actively pursued in the field of underwater environments [30]. In this regard, researchers have made significant contributions to the development of target detection algorithms. For instance, in 2017, Zhou et al. [31] integrated image enhancement techniques into an expanded VGG16 feature extraction network and employed a Faster R-CNN network with feature mapping for the detection and identification of underwater biological targets using the URPC dataset. In 2020, Chen et al. [32] introduced a new sample distribution-based weighted loss function called IMA (Invert Multi-Class AdaBoost) to mitigate the adverse effect of noise on detection performance. In 2021, Qiao et al. [33] proposed a real-time and accurate underwater target classifier, leveraging the combination of LWAP (Local Wavelet Acoustic Pattern) and MLP (Multilayer Perceptron) neural networks, to tackle the challenging problem of underwater target classification. Nevertheless, the joint requirement of localization and classification, in addition to classification, makes the target detection task especially challenging in underwater environments where images are often plagued by severe color distortion and low visibility caused by mobile acquisition.With the aim of enhancing the accuracy, achieving real-time performance, and promoting the portability of the underwater target detection capability, the most advanced YOLOv7 model of the YOLO series has been selected for improvement, resulting in the proposed YOLOv7-AC model, designed to address the difficulties encountered in this field. The effectiveness of the proposed model has been demonstrated through experiments conducted on underwater images.The innovations of this paper are as follows:

(1) In order to extract more informative features, the integration of the Global Attention Mechanism (GAM) [39] is proposed. This mechanism effectively captures both the channel and spatial aspects of the features and increases the significance of cross-dimensional interactions.

(2) To further enhance the performance of the network, the ACmix (A mixed model incorporating the benefits of self-Attention and Convolution) [40] is introduced.

(3) The design of the ResNet-ACmix module in YOLOv7-AC aims to enhance the feature extraction capability of the backbone network and to accelerate the convergence of the network by capturing more informative features.

(4) The E-ELAN module in the YOLOv7 network is optimized by incorporating Skip Connections and a 1x1 convolutional structure between modules and replacing the 3x3 Convolutional layer with the ACmixBlock. This results in an enhanced feature extraction ability and improved speed during inference.

The rest of this paper is organized as follows. Section 2 describes the architecture of YOLOv7 model and related methods. Section 3 presents the proposed YOLOv7-AC model and its theoretical foundations. The performance of the YOLOv7-AC model is evaluated and analyzed through experiments conducted on underwater image datasets in Section 4. The limitations and drawbacks of the proposed method are discussed in Section 5. Finally, we provide a conclusion of this work in Section 6.

## 2. Related Works

### 2.1. YOLOv7

The YOLOv7 model, developed by Chien-Yao Wang and Alexey Bochkovskiy et al. in 2022, integrates strategies such as E-ELAN (Extended efficient layer aggregation networks) [34], model scaling for concatenation-based models [35], and model re-parameterization [36] to achieve a favorable balance between detection efficiency and precision. As shown in Figure 1, the YOLOv7 network consists of distinct four modules: the Input module, the Backbone network, the Head network and the Prediction network.

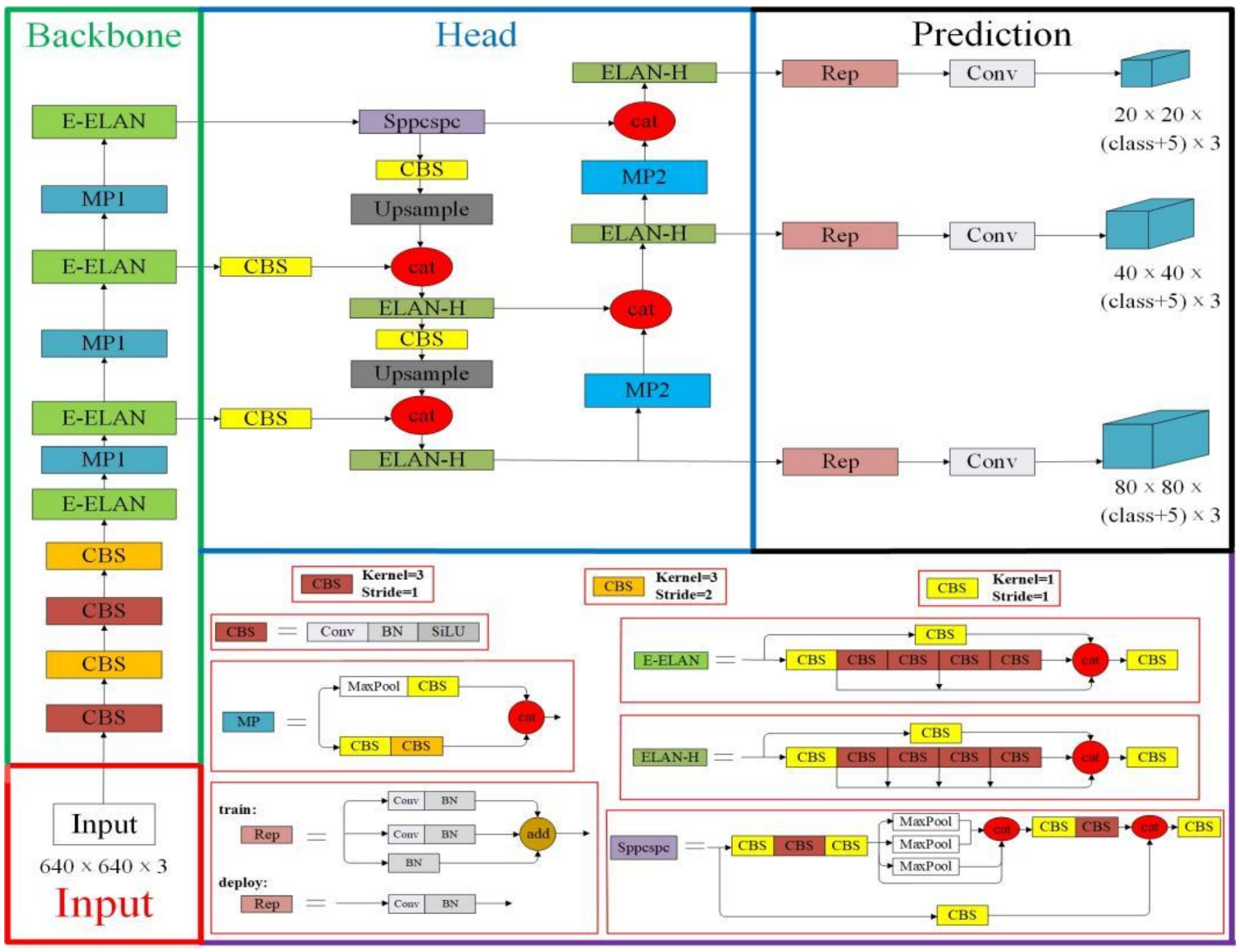

**Figure 1.** The network structure of YOLOv7 [29].

Input module: The pre-processing stage of the YOLOv7 model employs both mosaic and hybrid data enhancement techniques and leverages the adaptive anchor frame calculation method established by YOLOv5 to ensure that the input color images are uniformly scaled to a 640x640 size, thereby meeting the requirements for the input size of the backbone network.

Backbone network: The YOLOv7 network comprises three main components: CBS, E-ELAN, and MP1. The CBS module is composed of convolution, batch normalization, and SiLU activation functions. The E-ELAN module maintains the original ELAN design architecture and enhances the network's learning ability by guiding different feature group computational blocks to learn more diverse features, preserving the original gradient path. MP1 is composed of CBS and MaxPool and is divided into upper and lower branches. The upper branch uses MaxPool to halve the image's length and width and CBS with 128 output channels to halve the image channels. The lower branch halves the image channels through a CBS with a 1x1 kernel and stride, halves the image length and width with a CBS of 3x3 kernel and 2x2 stride, and finally fuses the features extracted from both branches through the concatenation (Cat) operation. MaxPool extracts the maximum value information of small local areas while CBS extracts all value information of small local areas, thereby improving the network's feature extraction ability.

Head network: The Head network of YOLOv7 is structured using the Feature Pyramid Network (FPN) architecture, which employs the PANet design. This network comprises several Convolutional, Batch Normalization and SiLU activation (CBS) blocks, along with the introduction of a Spatial Pyramid Pooling and Convolutional Spatial Pyramid Pooling (Sppcspc) structure, the extended efficient layer aggregation network (E-ELAN), and MaxPool-2 (MP2). The Sppcspc structure improves the network's perceptual field through the incorporation of a Convolutional Spatial Pyramid (CSP) structure within the Spatial Pyramid Pooling (SPP) structure, along with a large residual edge to aid optimization and feature extraction. The ELAN-H layer, which is a fusion of several feature layers based on E-ELAN, further enhances feature extraction. The MP2 block has a similar structure to the MP1 block, with a slight modification to the number of output channels.

Prediction network: The Prediction network of YOLOv7 employs a Rep structure to adjust the number of image channels for the features output from the head network, followed by the application of 1x1 convolution for the prediction of confidence, category, and anchor frame. The Rep structure, inspired by RepVGG [37], introduces a special residual design to aid in the training process. This unique residual structure can be reduced to a simple convolution in practical predictions, resulting in a decrease in network complexity without sacrificing its predictive performance.

### *2.2. GAM*

The attention mechanism is a method used to improve the feature extraction in complex contexts by assigning different weights to the various parts of the input in the neural network. This approach enables the model to focus on the relevant information and ignore the irrelevant information, resulting in improved performance. Examples of attention mechanisms include pixel attention, channel attention, and multi-order attention [38].

GAM [39] could improve the performance of deep neural networks by reducing information dispersion and amplifying the global interaction representation. The module structure is shown in Figure 2.

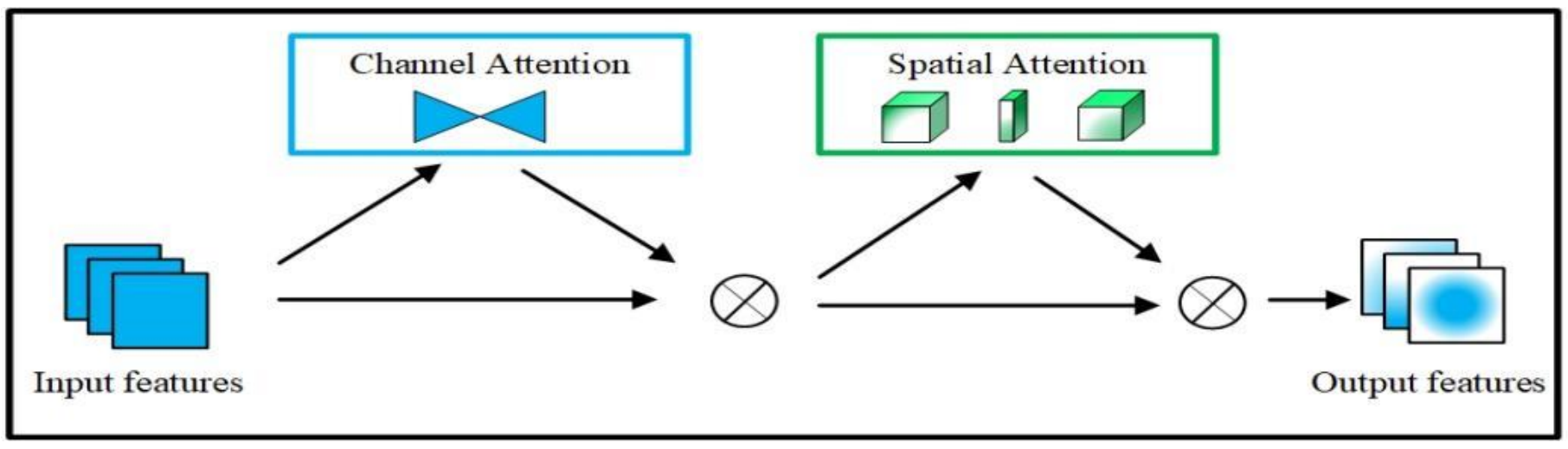

**Figure 2.** The structure diagram of GAM module [39].

The GAM encompasses a channel attention submodule and a spatial attention submodule. The channel attention submodule is designed as a three-dimensional transformation, allowing it to preserve the three-dimensional information of the input. This is followed by a multi-layer perception (MLP) with two layers, which serves to amplify the inter-dimensional dependence in the channel space, thus enabling the network to focus on the more meaningful and foreground regions of the image, as depicted in Figure 3.

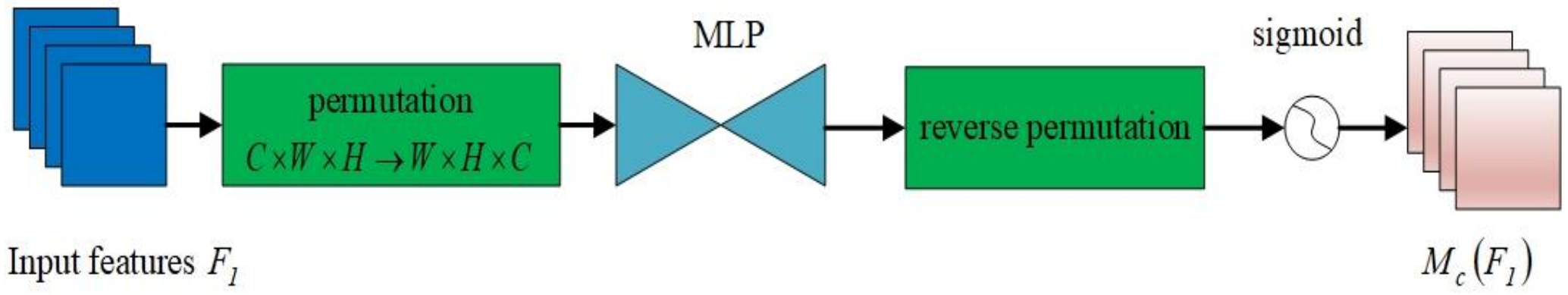

**Figure 3.** The structure diagram of channel attention submodule in GAM module [39].

The spatial attention submodule incorporates two convolutional layers to effectively integrate spatial information, enabling the network to concentrate on contextually significant regions across the image, as depicted in Figure 4.

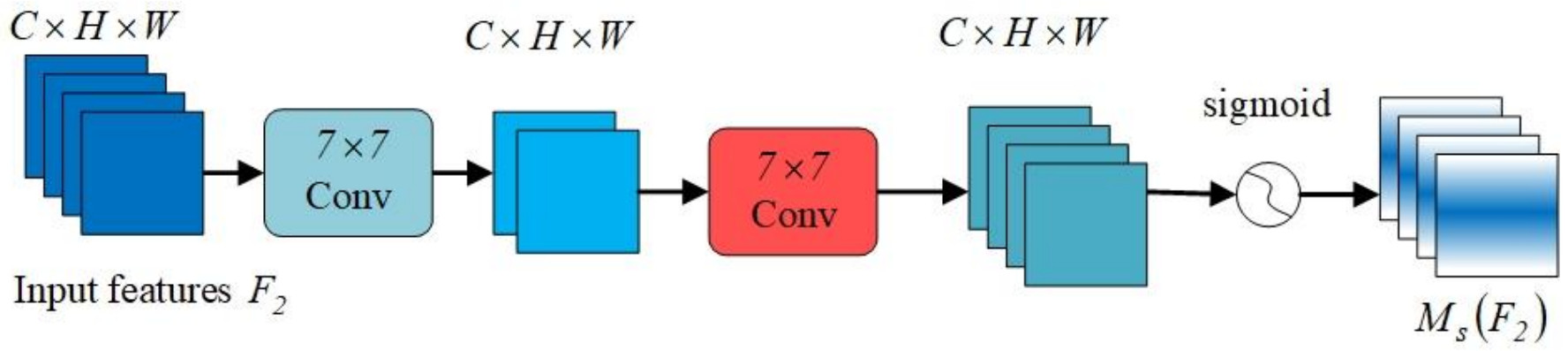

**Figure 4.** The structure diagram of spatial attention submodule in GAM module [39].

*2.3. ACmix*

The authors of [40] discovered that self-attention and convolution both heavily rely on the 1x1 convolution operation. To address this, they developed a hybrid model known as ACmix, which elegantly combines self-attention and convolution with minimal computational overhead. The model structure is illustrated in Figure 5.

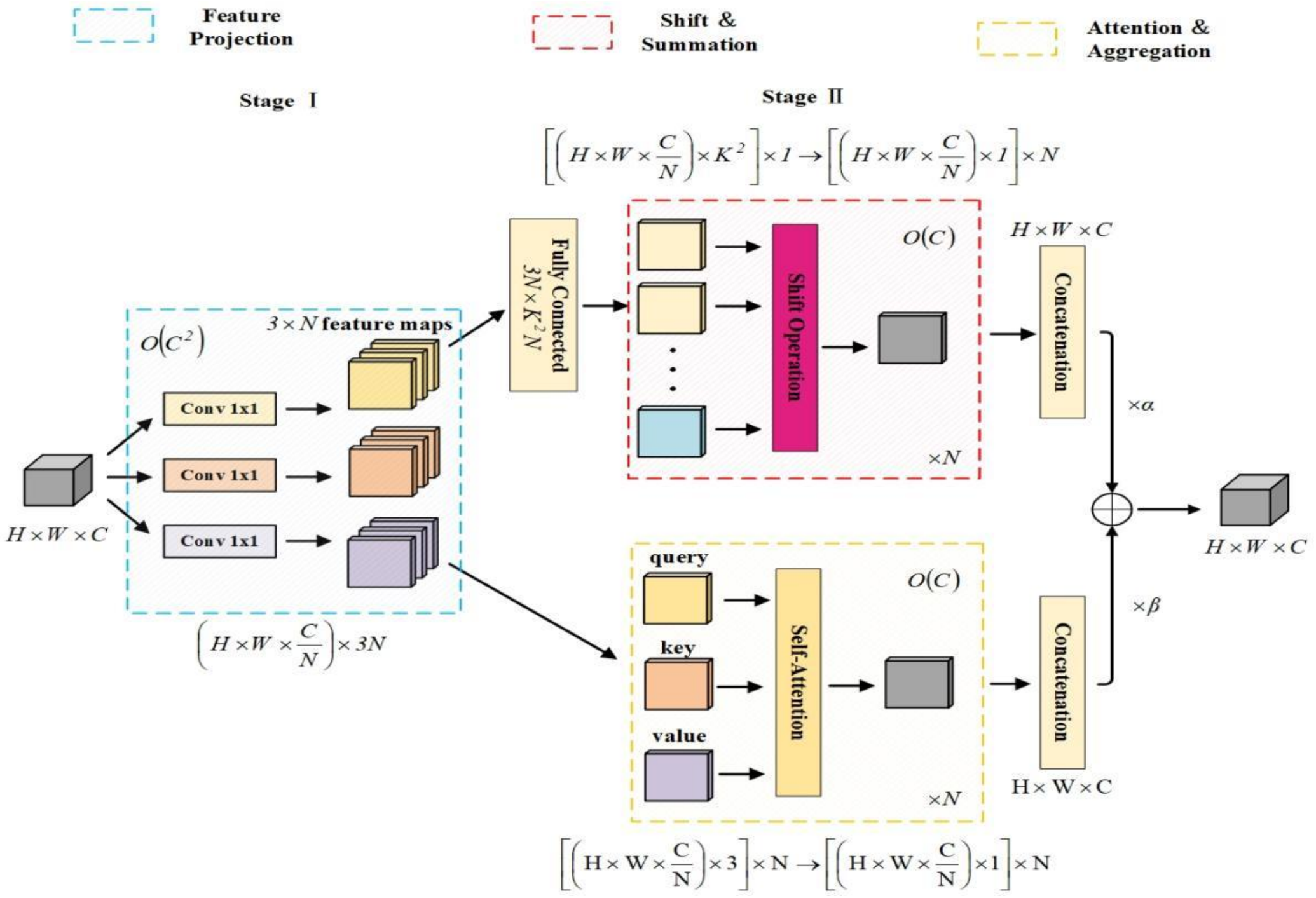

**Figure 5.** The structure diagram of ACmix module [40].

The first core is convolution [41]: Given a standard convolution of the kernel $K \in \Re^{C_{out} \times C_{in} \times k \times k}$, tensor $F \in \Re^{C_{in} \times H \times W}$, $G \in \Re^{C_{out} \times H \times W}$ as input and output feature maps, respectively, $k$ is the kernel size, $C_{in}$ and $C_{out}$ are input and output channels. $H$, $W$ denote height and width. $f_{ij} \in \Re^{C_{in}}$, $g_{ij} \in \Re^{C_{out}}$ denote the feature tensor of the pixels $(i, j)$ corresponding to $F$ and $G$. The standard convolution can be expressed as equation (1):

$$g_{i,j} = \sum_{p,q} K_{p,q} f_{i+p-[k/2],\, j+q-[k/2]} \tag{1}$$

where $K_{p,q} \in \Re^{C_{out} \times C_{in}}$ subjecting to $p, q \in \{0, 1, ..., k-1\}$ denotes the weight of the nucleus at position $(p, q)$. Equation (1) can be simplified into equations (2) and (3).

$$g_{ij} = \sum_{p,q} g_{ij}^{(p,q)} \tag{2}$$

$$g_{ij}^{(p,q)} = K_{p,q} f_{i+p-[k/2],\, j+q-[k/2]} \tag{3}$$

To further simplify the formula, define the Shift operation by $\tilde{f} \triangleq Shift(f, \Delta x, \Delta y)$

$$\tilde{f}_{i,j} = f_{i+\Delta x,\, j+\Delta y}, \forall i, j \tag{4}$$

where $\Delta x, \Delta y$ correspond to horizontal and vertical displacements. Thus, equation (3) can be abbreviated to

$$g_{ij}^{(p,q)} = Shift\left(K_{p,q} f_{ij}, p-[k/2], q-[k/2]\right) \tag{5}$$

Standard convolution can be summarized in two steps:
(Convolution:)

State Ⅰ: $$\tilde{g}_{ij}^{(p,q)} = K_{p,q} f_{ij} \tag{6}$$

State Ⅱ: $$g_{ij}^{(p,q)} = Shift\left(\tilde{g}_{ij}^{(p,q)}, p-[k/2], q-[k/2]\right) \tag{7}$$

$$g_{ij} = \sum_{p,q} g_{ij}^{(p,q)} \tag{8}$$

The next is Self-Attention [42]: Assuming that there is a standard self-attentive module with $N$ heads, the output of the attention module is calculated as:

$$g_{ij} = \prod_{l=1}^{N} \left( \sum_{a,b \in E_k(i,j)} A\left(W_q^{(l)} f_{ij}, W_k^{(l)} f_{ab}\right) W_v^{(l)} f_{ab} \right) \tag{9}$$

where $\Pi$ is the concatenation of the output of $N$ attention headers, $W_q^{(l)}$, $W_k^{(l)}$, $W_v^{(l)}$ are the projection matrices of query, key and value. $E_k(i, j)$ denotes a local region of pixels of spatial width $k$ centered on $(i, j)$. $A\left(W_q^{(l)} f_{ij}, W_k^{(l)} f_{ab}\right)$ is the corresponding attention weight with regard to the features within $E_k(i, j)$. For the widely adopted self-attentive modules, the $A\left(W_q^{(l)} f_{ij}, W_k^{(l)} f_{ab}\right)$ weights are calculated as:

$$A\left(W_q^{(l)} f_{ij}, W_k^{(l)} f_{ab}\right) = softmax_{E_k(i,j)} \left( \frac{\left(W_q^{(l)} f_{ij}\right)^T \left(W_k^{(l)} f_{ab}\right)}{\sqrt{d}} \right) \tag{10}$$

where $d$ is the characteristic dimension of $W_q^{(l)} f_{ij}$. Similarly, multi-head self-attention can be decomposed into two stages:
(Self-Attention:)

State Ⅰ: $$q_{ij}^{(l)} = W_q^{(l)} f_{ij}, k_{ij}^{(l)} = W_k^{(l)} f_{ij}, v_{ij}^{(l)} = W_v^{(l)} f_{ij} \tag{11}$$

State Ⅱ: $$g_{ij} = \prod_{l=1}^{N} \left( \sum_{a,b \in E_k(i,j)} A\left(q_{ij}^{(l)}, k_{ab}^{(l)}\right) v_{ab}^{(l)} \right) \tag{12}$$

The ACmix module performs the 1x1 convolution projection operation on the input feature mapping once and reuses the intermediate feature maps for subsequent aggregation operations in both Convolution and self-attention paths. The strengths of these outputs are controlled by two learnable scalars:

$$F_{out} = \alpha F_{att} + \beta F_{conv} \tag{13}$$

The ACmix module reveals the robust correlation between self-attention and convolution, leveraging the advantages of both techniques, and mitigating the need for repeated, high-complexity projection operations. As a result, it offers minimal computational overhead compared to either self-attention or pure convolution alone.

## 3. Method

This section outlines the design of an underwater target detection model that leverages the improved YOLOv7 architecture. We design a ResNet-ACmix module (Section 3.1) and AC-E-ELAN structure (Section 3.2), finally resulting in the improved YOLOv7-AC model (Section 3.3).

### *3.1. ResNet-ACmix module*

The introduction of the ResNet-ACmix module into the Backbone component of YOLOv7 effectively preserves the coherence of the extracted feature information. This module is based on the bottleneck structure of ResNet [43], wherein the 3x3 convolution is substituted by the ACmix module, enabling adaptive focus on different regions and capturing more informative features, as illustrated in Figure 6. The input is divided into the main input and residual input, which helps prevent information loss, while reducing the number of parameters and computational requirements. The ResNet-ACmix module enables the network to attain deeper depths without encountering gradient disappearance, and the learning outcomes are more sensitive to fluctuations in network weights.

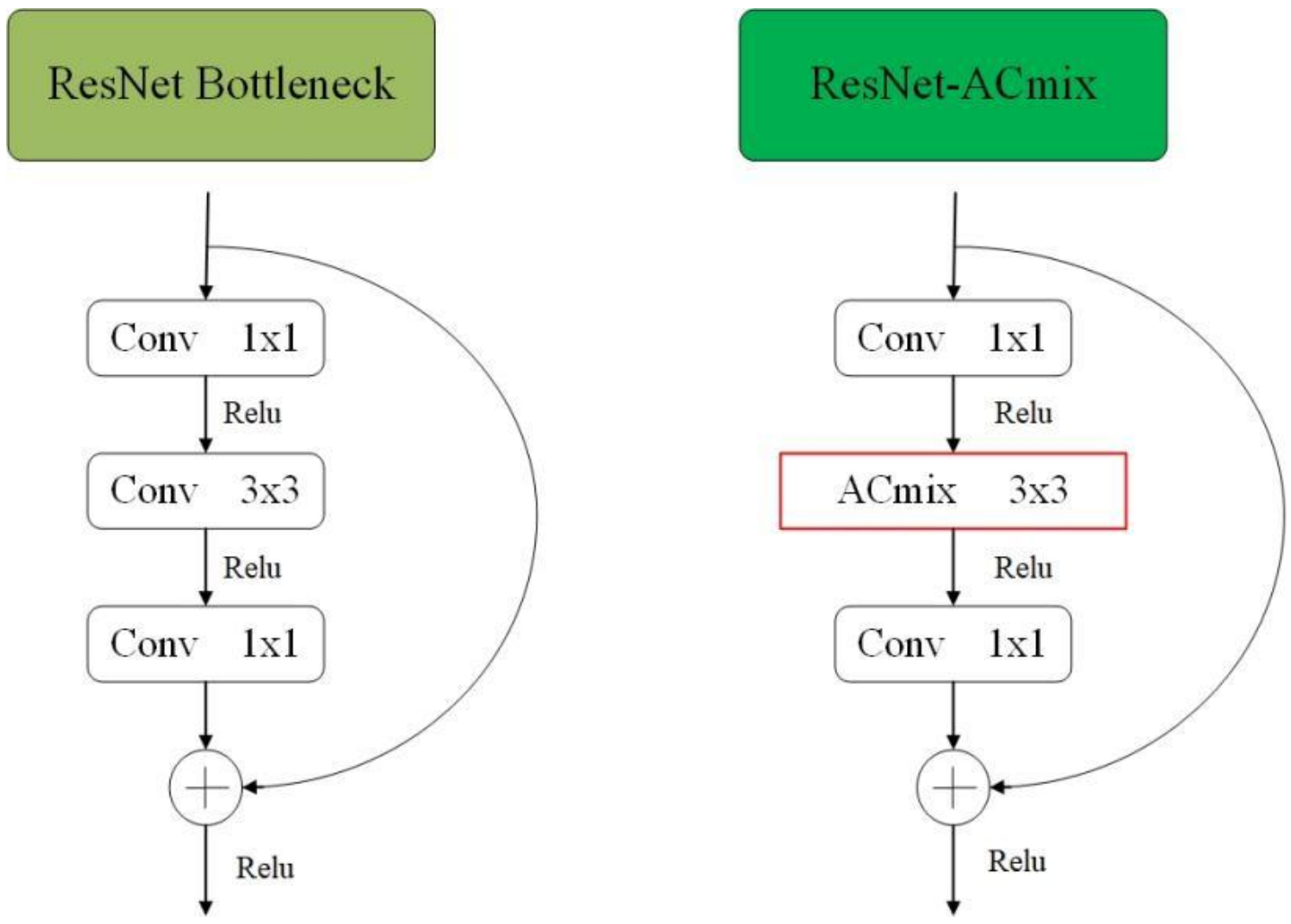

**Figure 6.** The structure diagram of ResNet-ACmix module (**left:** ResNet; **right:** ResNet-ACmix).

### *3.2. AC-E-ELAN module*

The proposed improvement to the E-ELAN component of YOLOv7 is based on the advanced ELAN architecture [44]. Unlike traditional ELAN networks, the extended E-ELAN employs an expand, shuffle, and merge cardinality approach that enables continuous enhancement of the network's learning capability without disrupting the original gradient flow, thereby enhancing parameter utilization and computational efficiency. The feature extraction module of the E-ELAN component in YOLOv7 has been further improved by incorporating residual structures (i.e., 1x1 convolutional branch and jump connection branch) from the RepVgg architecture [45]. This has led to the development of the AC-E-ELAN structure, as depicted in Figure 7, which integrates the ACmixBlock, consisting of 3x3 convolutional blocks, with jump connections and 1x1 convolutional structures between the ACmixBlocks. This combination enables the network to benefit from both the rich features obtained during the training of a multi-branch model and the fast, memory-efficient inference obtained from a single-path model.

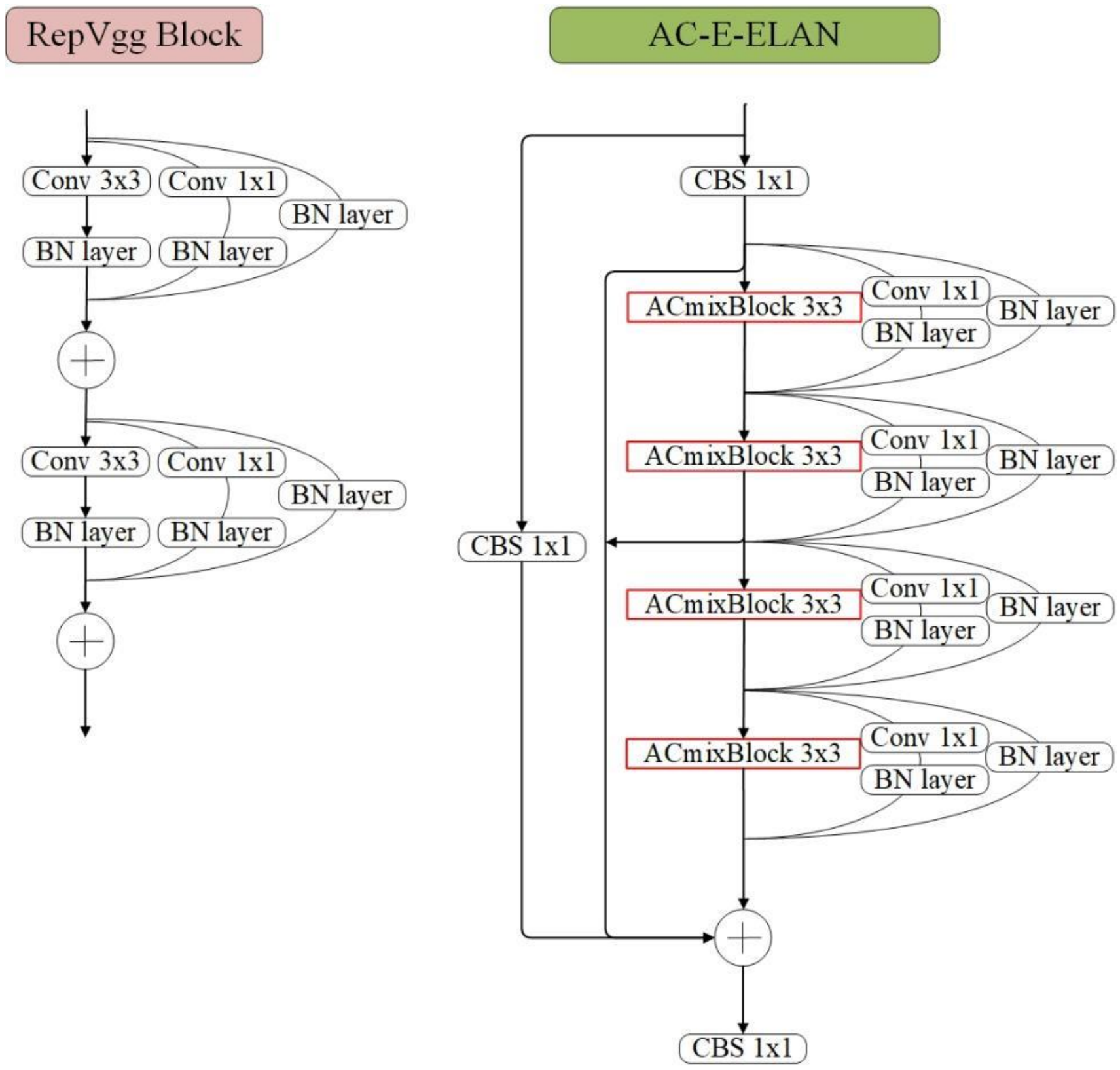

**Figure 7.** The structure diagram of AC-E-ELAN module (**left:** RepVgg; **right:** AC-E-ELAN).

### *3.3. The proposed YOLOv7-AC model*

In the proposed YOLOv7-AC model, the original E-ELAN network in YOLOv7 is improved by the introduction of the AC-E-ELAN structure. The 3x3 convolutional blocks are replaced with 3x3 ACmixBlock, and jump connections and 1x1 convolutional structures are added between the ACmixBlock blocks, enhancing the ability of the model to focus on the valuable content and location of the input image samples, enriching the features extracted by the network, and reducing the model's inference time. Additionally, the ResNet-ACmix blocks are integrated into the Backbone module, located behind the fourth CBS and at the bottom layer of the Backbone, to effectively retain the features collected by the Backbone and extract feature information of small targets and complex background targets, while simultaneously accelerating the network's convergence and improving the detection accuracy.

The incorporation of the GAM in the Backbone and Head of YOLOv7 enhances the network's ability to extract deep and important features effectively.

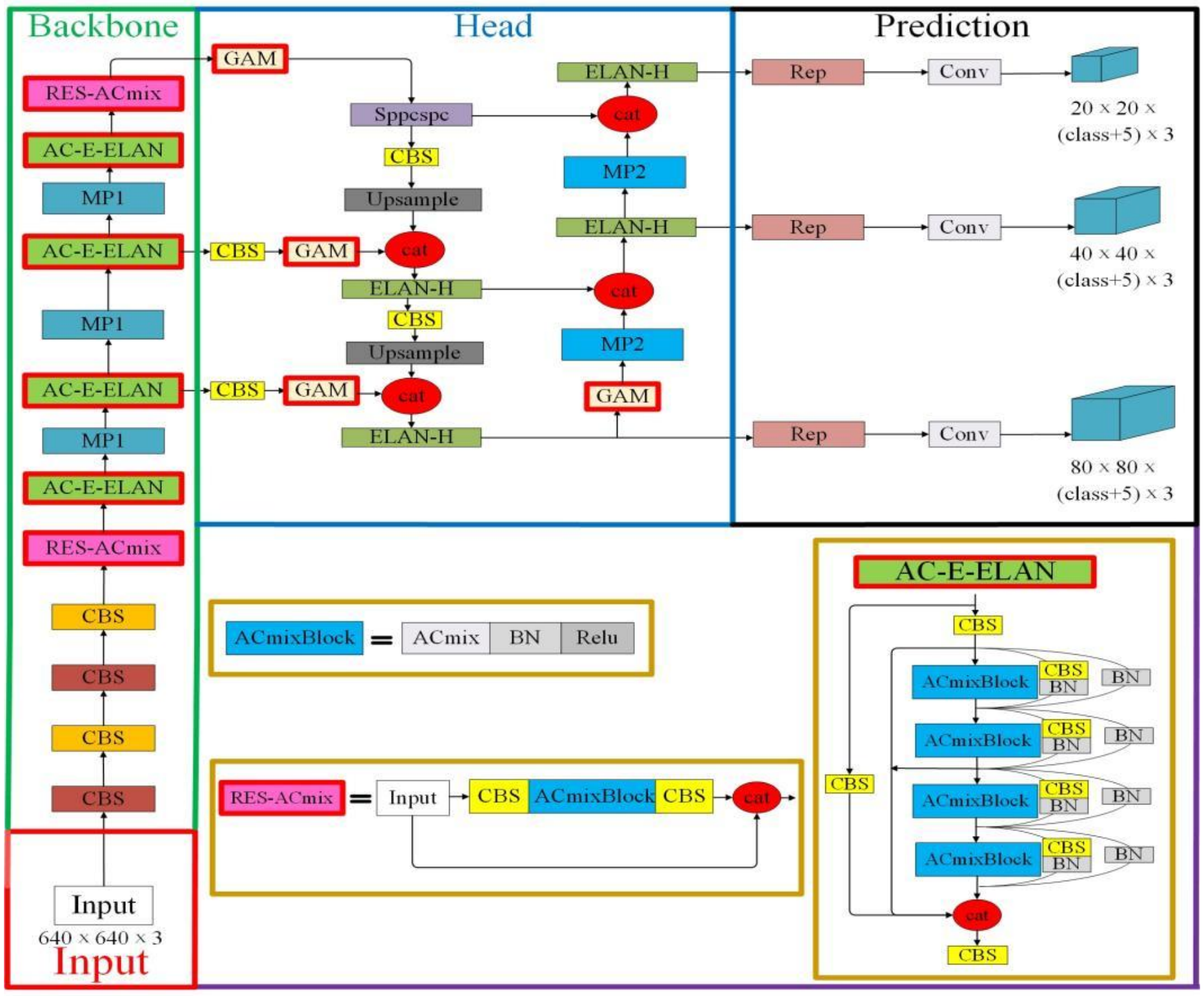

**Figure 8.** Structure diagram of the YOLOv7-AC model.

## 4. Experiments

In this section, the configuration of the experimental environment, hyperparameters, test dataset, and optimization of anchor boxes are described. The experimental results demonstrate that the proposed YOLOv7-AC model enhances both accuracy and speed in underwater target detection, thereby verifying its effectiveness and superiority in the challenging underwater detection environment.

### *4.1. Experimental environment*

The experimental platform is equipped with 5 vCPU Intel(R) Xeon(R) Silver 4210 CPU @ 2.20GHz, NVIDIA GeForce RTX 3090 GPU with 24GB video memory size, Windows 11, 64-bit operating system. The software environments are CUDA 11.3, CUDNN 8.2.2, and the compiler Python 3.9.

### *4.2. Model hyperparameter setting*

The effectiveness of the YOLOv7-AC model was evaluated by training and testing the neural network using the hyperparameters detailed in Table 1.

**Table 1.** Experimental configuration.

| Parameter | Configuration |
|---|---|
| learning rate | 0.01 |
| momentum | 0.937 |
| weight decay | 0.0005 |
| batch size | 16 |
| optimizer | SGD |
| image size | 640x640 |
| epochs | 300 |

*4.3. The URPC Dataset*

This dataset was the 2021 National Underwater Robot Professional Competition (URPC) dataset, with 7600 underwater optical images (including manually annotated truth data). For more information, please find https://www.heywhale.com/home/competition/605ab78821e3f6003b56a7d8/content/0. The target groups to be evaluated in the experiment consist of four categories of seafood, namely "holothurian", "echinus", "scallop", and "starfish". In the traditional seafloor target detection often focus on sea food, so this paper remove the seagrass-related samples from the dataset, and keep 6575 valid images. Some of example images in the URPC dataset are shown in Figure 9.

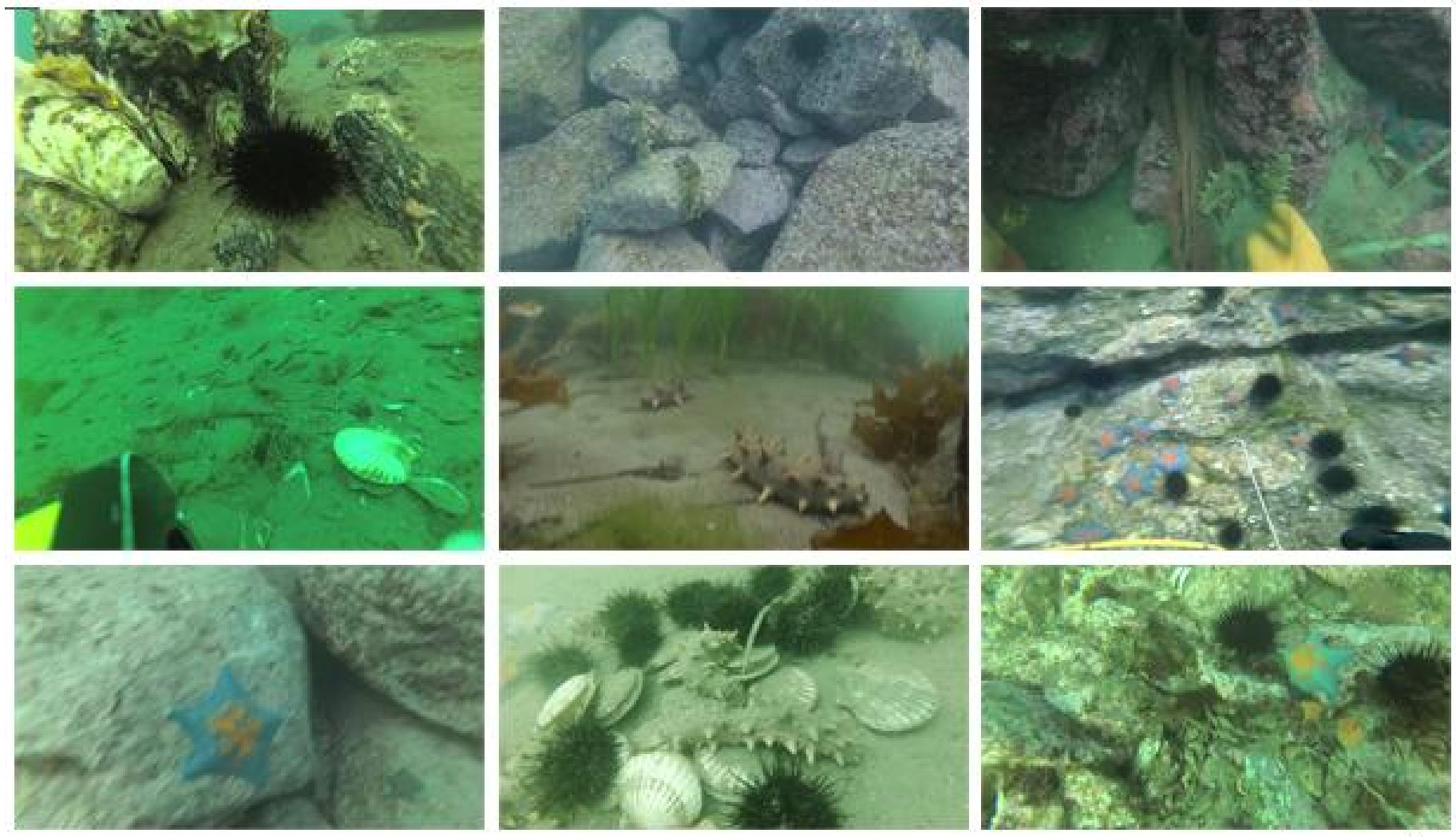

**Figure 9.** Example images of the URPC dataset.

As depicted in Figure 10(a), the number of targets in each category was analyzed, with the most abundant being the echinus category, followed by starfish, scallop, and holothurian. The regularized target location map, as shown in Figure 10(b), reveals that the targets are more densely concentrated in the horizontal direction and comparatively dispersed in the vertical direction. Additionally, the normalized target size maps in Figure 10(c) indicate that the target sizes are relatively concentrated, with a majority of them being small. To create the experimental dataset, a 7:3 ratio of training set to test set was established, with 4521 images comprising the training set and 2054 images comprising the test set, divided randomly.

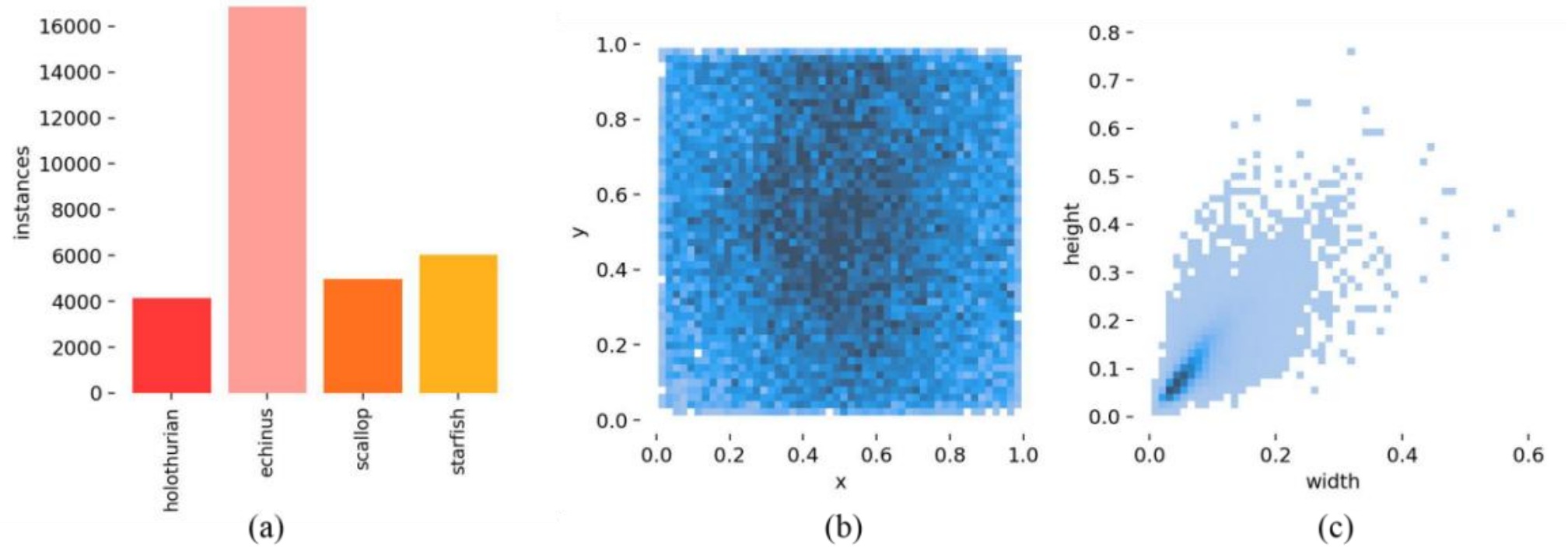

**Figure 10.** Statistical results of the URPC dataset: (**a**) bar chart of the number of targets in each class; (**b**) normalized target location map; (**c**) normalized target size map.

4.3.1. K-means++ to get anchor box of the URPC dataset

In order to enhance the accuracy and efficiency of detection, this study employs the K-means++ algorithm, instead of the traditional K-means algorithm, to cluster the anchor boxes of the URPC dataset. The K-means++ algorithm selects the initial clustering centers based on the principle that the distance between the centers should be maximized, which is an improvement over the traditional K-means algorithm that randomly selects k data objects as initial centers. Additionally, the distance indicator used in the clustering process is changed from the Euclidean distance to 1-IOU (boxes, anchors). The YOLOv7 model has three detection feature maps, with each feature map corresponding to three anchors, resulting in a total of nine anchors. The dimensions of the three feature maps and the corresponding anchor boxes are specified in Table 2.

**Table 2.** Anchor box parameters of the URPC dataset.

| Feature Map Size | $80\times80$(px) | $40\times40$(px) | $20\times20$(px) |
|---|---|---|---|
| | 28,25 | 68,65 | 182,160 |
| YOLOv7 | 39,37 | 96,84 | 272,219 |
| | 55,46 | 139,112 | 436,362 |

4.3.2. Model evaluation metrics

Commonly used basic metrics for target detection are Precision, Recall, Intersection over Union (IOU), etc. AP (Average - Precision) and mAP (mean Average Precision) value. The IOU metric is employed to measure the degree of overlap between the system-predicted bounding box and the ground-truth bounding box in the original image. ,The calculation is the intersection and concatenation ratio of Detection Result to Ground Truth as follow.

$$IOU=\frac{DetectionResult\cap GroudTruth}{DetectionResult\cup GroudTruth} \tag{14}$$

In the experiment, a threshold value of IOU was established. The Detection Result was considered as True Positive (TP) when the IOU value calculated between the Detection Result and the Ground Truth was greater than the threshold value, indicating the accurate identification of targets. Conversely, the Detection Result was considered as False Positive (FP) when the IOU value was less than the threshold value, representing incorrect identification. The number of undetected targets, referred to as False Negative (FN), was calculated as the number of Ground Truth instances with no matching Detection Result.

Precision is defined as the proportion of True Positives in the recognized images, expressed as a percentage.

$$Precision = \frac{TP}{TP + FP} \quad (15)$$

Recall is the proportion of all positive samples in the test set that identify the correct target.

$$Recall = \frac{TP}{TP + FN} \quad (16)$$

A Precision-Recall (PR) curve plots precision along the vertical axis and recall along the horizontal axis, thereby illuminating the interplay between the accuracy of the classifier in identifying positive instances and its ability to encompass all positive instances. The Average Precision (AP) is a scalar representation of the area under the PR curve, with higher values indicating superior performance of the classifier.

$$AP = \int_0^1 Precision(Recall)\, dRecall \quad (17)$$

In target detection, the model usually detects many classes of targets, and each class can plot a PR curve, which in turn calculates an AP value. mAP represents the average of the APs of all classes.

$$mAP = \frac{1}{class_nunber} \sum_{1}^{class_number} AP \quad (18)$$

4.3.3. Experimental results and analysis of the URPC dataset

The proposed YOLOv7-AC model was experimentally evaluated on the URPC dataset with respect to its detection performance. As illustrated in Figure 11, the results of the improved model demonstrate an improved detection efficiency on the various target categories, particularly the echinus category which boasts an average precision (AP) value of 92.2%. The mean average precision (mAP) for the model was calculated to have a value of 89.6%.

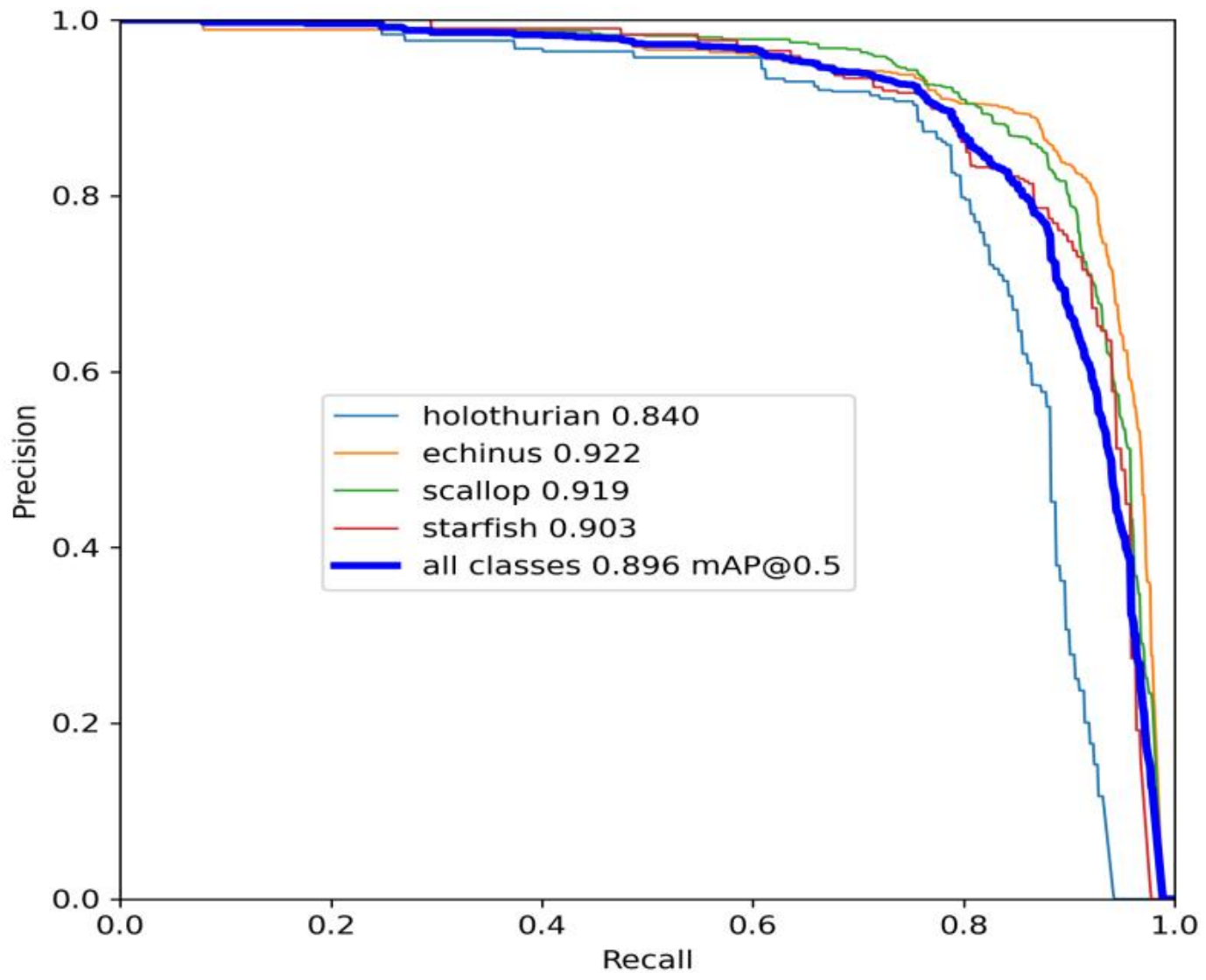

**Figure 11.** The precision-recall curve of the YOLOv7-AC model on the URPC dataset.

A confusion matrix was utilized to evaluate the accuracy of the proposed YOLOv7-AC model's results. Each column of the confusion matrix represents the predicted proportions of each category, while each row represents the true proportions of the respective category in the data, as depicted in Figure 12. The analysis of Figure 12 reveals that the predicted categories of "holothurian", "echinus", "scallop" and "starfish" have correct prediction rates of 81%, 90%, 89% and 89%, respectively, which suggests that the model has a high degree of accuracy.

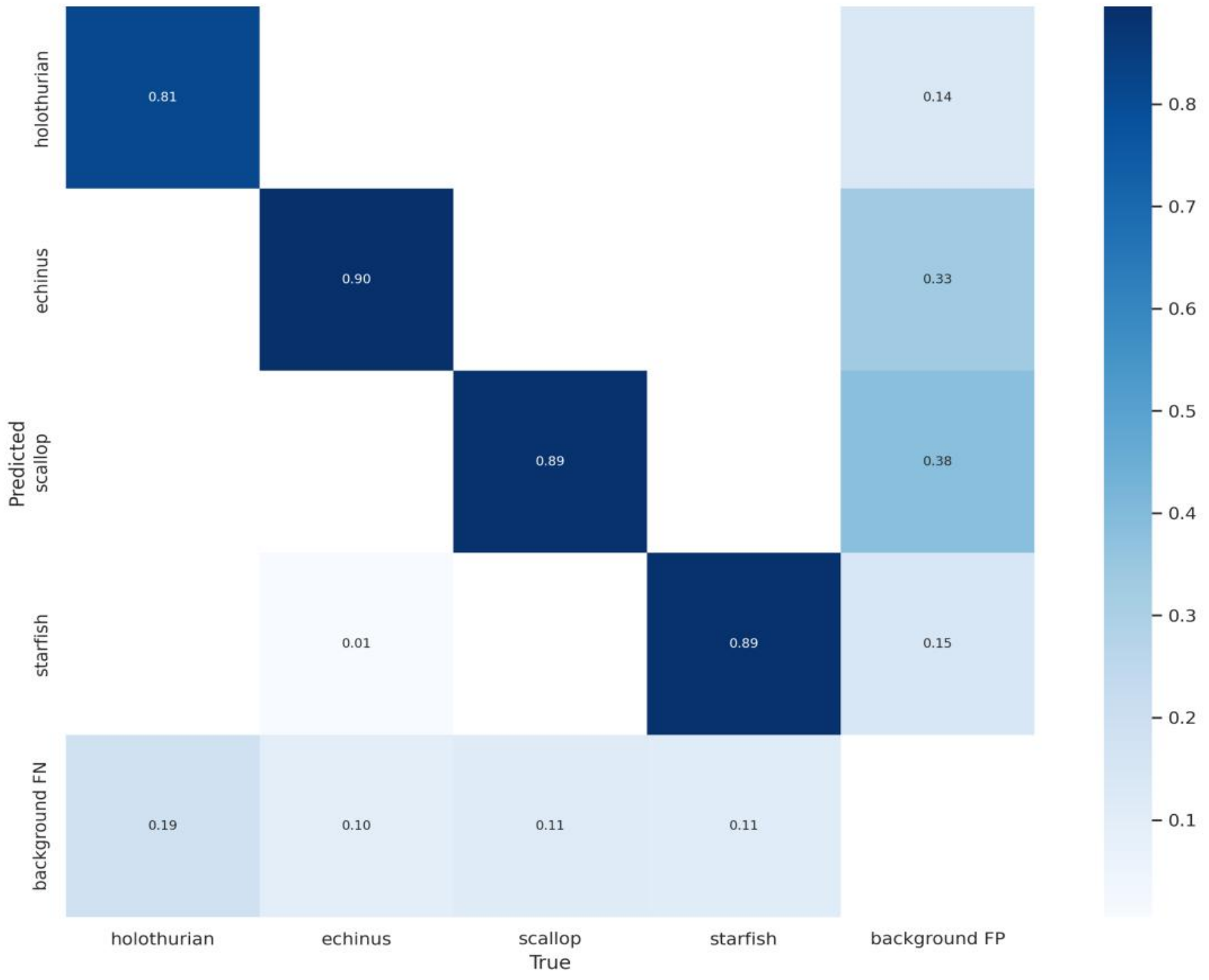

**Figure 12.** The confusion matrix of the YOLOv7-AC model on the URPC dataset.

Additionally, this study presents the graphical representation of the variations in the loss values including the Box loss, Objectness loss, and Classification loss. YOLOv7 adopts the GIOU Loss as the loss function for bounding boxes, where the Box loss is the mean of the GIOU loss function and a lower value indicates higher accuracy. The Objectness loss is the average value of the target detection loss, with a smaller value corresponding to higher accuracy. The Classification loss is the mean of the classification loss, with a lower value indicating higher accuracy, as demonstrated in Figure 13. As illustrated in Figure 13, the loss values demonstrate a steady decrease and eventual stabilization as the number of iterations increases, reaching convergence after 200 iterations.

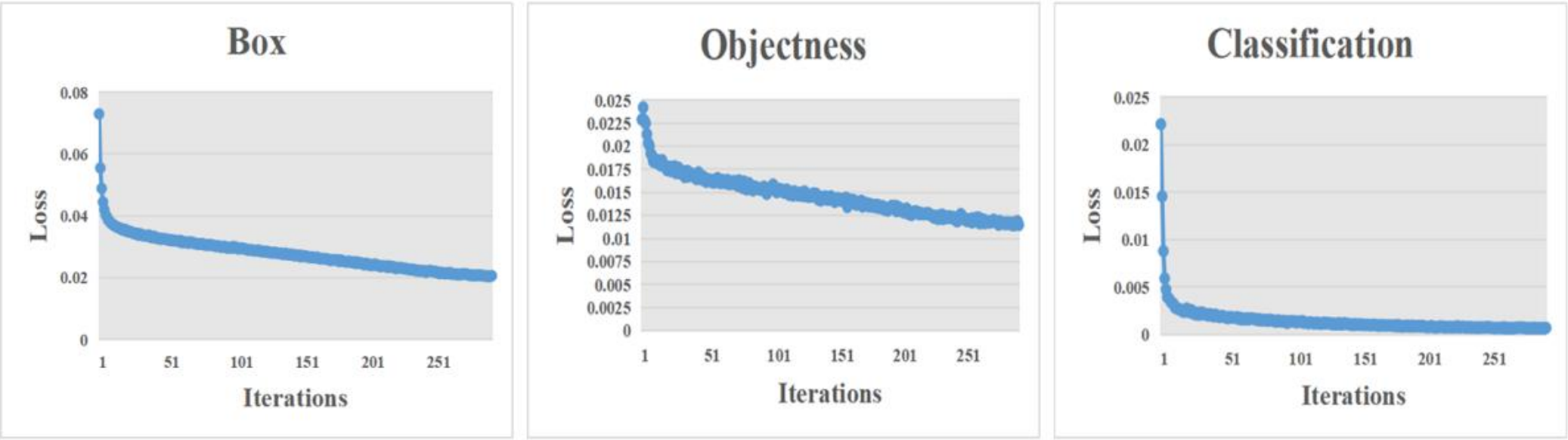

**Figure 13.** Variation curves of loss values on the URPC dataset.

To further demonstrate the superiority of the proposed YOLOv7-AC model, a comparison was performed with popular target detection models, including YOLOv7, YOLOv6, YOLOv5s, SSD, etc. by conducting training and testing on the URPC dataset and comparing their evaluation metrics, such as mean Average Precision (mAP). The results of the comparison are presented in Table 3. As can be seen from the table, the YOLOv7-AC model outperforms the other detection algorithms, with mAP 3.4% higher than that of YOLOv7 and 6.1%, 6.4%, and 14.2% higher than that of YOLOv6, YOLOv5s, and SSD, respectively. These experimental results demonstrate the practical advantages of the proposed method in underwater target recognition.

**Table 3.** Performance comparison of target detection model on the URPC dataset.

| Method | Precision | Recall | mAP@0.5 | mAP@0.95 |
|---|---|---|---|---|
| EfficientDet-d0[47] | 82.2% | 73.1% | 80.5% | 41.1% |
| SSD[21] | 74.2% | 68.7% | 75.4% | 38.8% |
| RetinaNet-50[48] | 75.2% | 66.6% | 73.25% | 32.5% |
| Detr[49] | 85.2% | 80.9% | 84.6% | 46.6% |
| YOLOv5s[27] | 85.9% | 78.3% | 83.2% | 49.6% |
| YOLOv6[28] | 87.7% | 79.1% | 83.5% | 50.8% |
| YOLOv7[29] | 85.9% | 82.2% | 86.2% | 52.1% |
| YOLOv7-AC | **90.0%** | **84.2%** | **89.6%** | **53.7%** |

4.3.4. Ablation experiments of the URPC dataset

We performed ablation experiments in order to assess the effectiveness of different improvements on the model performance. Firstly, this paper uses the designed ResNet-ACmix and AC-E-ELAN models to extract features, then introduces GAM, and finally applies K-means++ to obtain the anchor box of the URPC dataset in this paper. The experimental results are shown in Table 4.

**Table 4.** Ablation comparison of model performance improvement on the URPC dataset.

| Model | ResNet -ACmix | AC-E-ELAN | GAM | K-means++ | AP （echinus） | AP （starfish） | AP （scallop） | AP （holothurian） | mAP |
|---|---|---|---|---|---|---|---|---|---|
| | × | × | × | × | 90.9% | 87.0% | 90.3% | 76.8% | 86.2% |
| YOLOv7 | √ | × | × | × | 90.1% | 89.5% | 90.3% | 79.4% | 87.3% |
| | √ | √ | × | × | 91.9% | 89.5% | 91.7% | 83.4% | 89.1% |

| | | | | | | | | |
|---|---|---|---|---|---|---|---|---|
| √ | √ | √ | × | 91.2% | 90.4% | 91.6% | 83.9% | 89.3% |
| √ | √ | √ | √ | 92.2% | 90.3% | 91.9% | 84.0% | 89.6% |

As can be seen from Table 4, the use of the ResNet-ACmix module resulted in a 1.1% increase in mAP value, and the AC-E-ELAN module acting as the model's backbone network to obtain more useful features was the most critical improvement, which increased the model's mAP by another 1.8% on top of the introduction of ResNet-ACmix. Finally, by incorporating GAM and using K-means++ clustering anchor box, mAP is also improved by 0.2% and 0.3% respectively based on the pre-experiments.

*4.4. The Brackish Dataset*

The Brackish dataset [50] is the first publicly available European underwater image dataset with 2465 images, including "fish", "small_fish", "crab", "shrimp", "jellyfish", and "starfish". Some of example images in the dataset are shown in Figure 14.

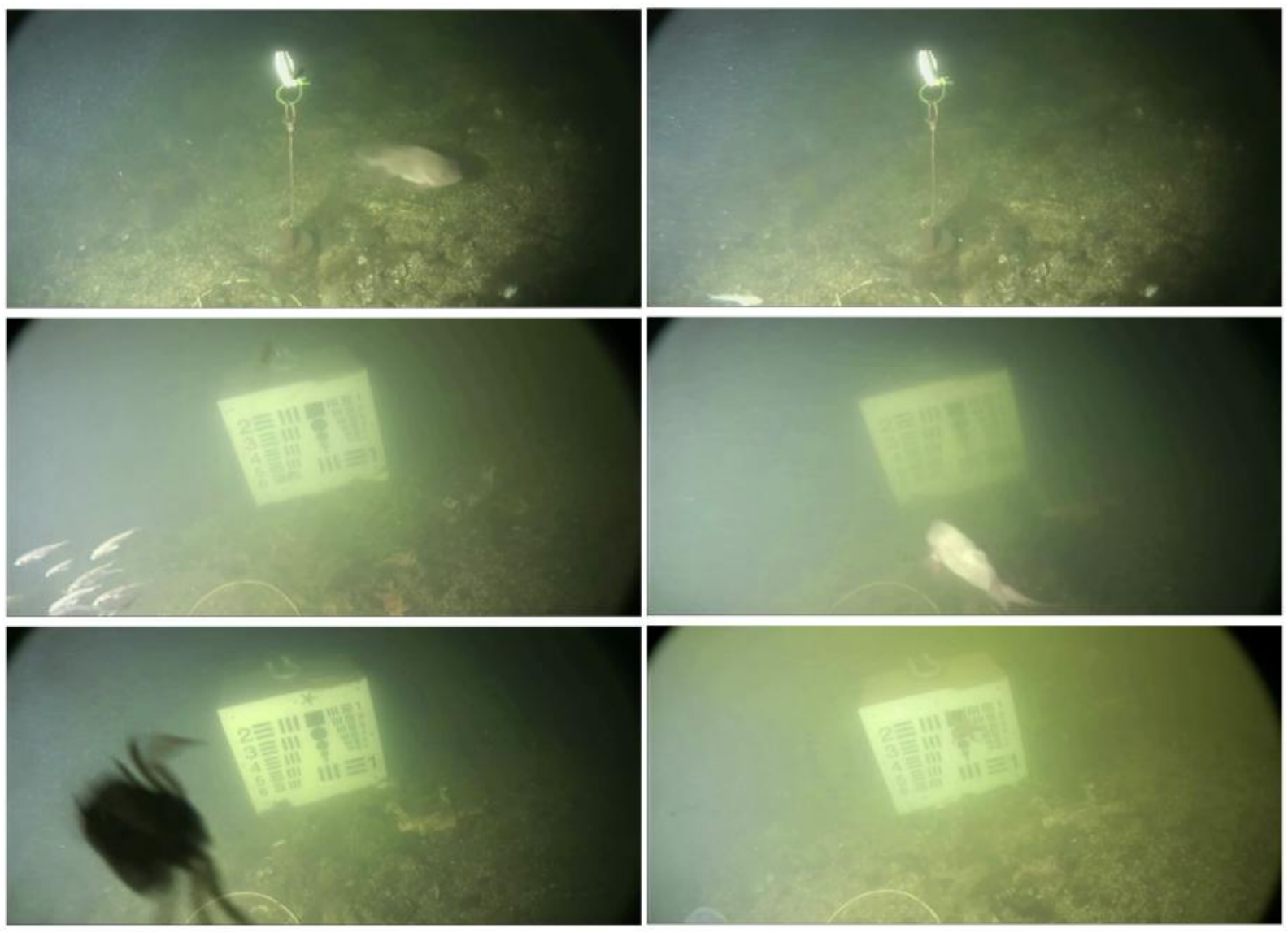

**Figure 14.** Example images of the Brackish dataset.

Figure 15(a) presents the distribution of the number of targets across various categories, with "Crab" having the largest representation. The regularized target location map is depicted in Figure 15(b), while Figure 15(c) illustrates the normalized target size maps, which reveal a majority of small targets. The URPC dataset was randomly divided into training and testing sets in a 7:3 ratio, with 1726 images being designated as the training set and 739 images as the testing set.

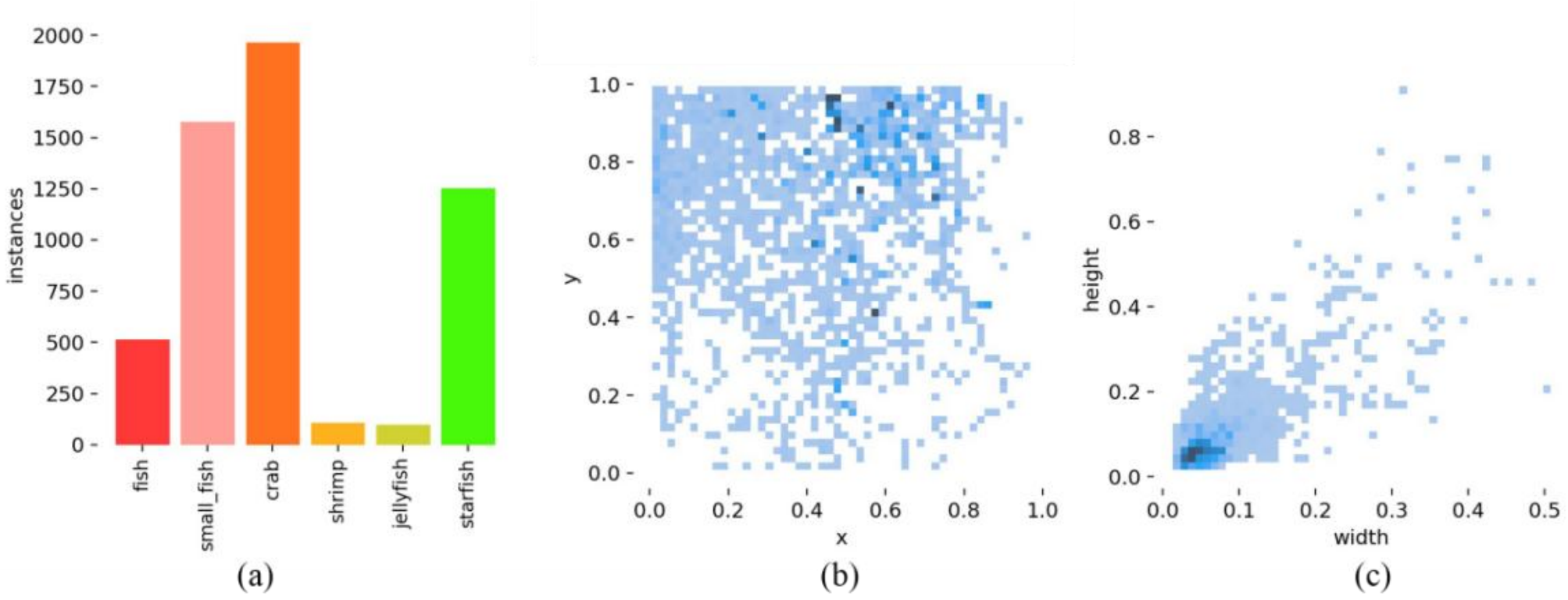

**Figure 15.** Statistical results of the Brackish dataset: (**a**) bar chart of the number of targets in each class; (**b**) normalized target location map; (**c**) normalized target size map.

4.4.1. K-means++ to get anchor box of the Brackish dataset

This study employs the K-means++ algorithm to cluster the anchor boxes of the Brackish dataset. The dimensions of the three feature maps and the corresponding anchor boxes are shown in Table 5.

**Table 5.** Anchor box parameters of the Brackish dataset.

| Feature Map Size | 80 × 80(px) | 40 × 40(px) | 20 × 20(px) |
|---|---|---|---|
| | 28,18 | 49,36 | 68,58 |
| YOLOv7 | 34,30 | 42,52 | 107,73 |
| | 53,20 | 73,35 | 222,170 |

4.4.2. Experimental results and analysis of the Brackish dataset

The effectiveness of the proposed YOLOv7-AC model was evaluated by conducting experiments on the Brackish dataset. The results of the detection performance are depicted in the precision-recall curve presented in Figure 16. As can be observed from the figure, the YOLOv7-AC model demonstrates improved performance in detecting various targets, particularly shrimp and starfish, which achieved an average precision value of 99.5%. The mAP of the model was 97.4%.

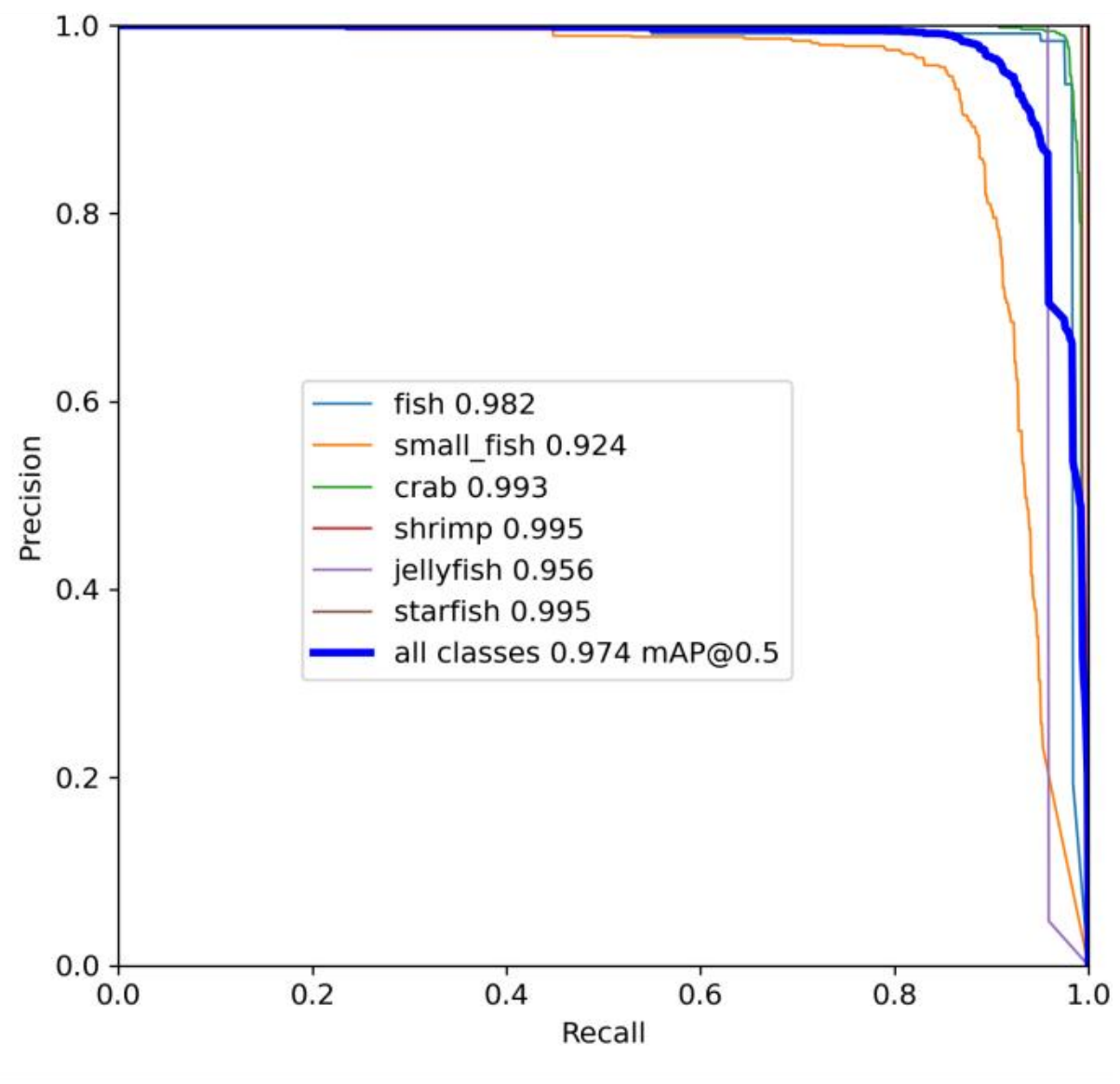

**Figure 16.** The precision-recall curve of the YOLOv7-AC model on the Brackish dataset.

The confusion matrix with regard to the Brackish dataset was shown in Figure 17. As can be seen from Figure 17, most of the targets were correctly predicted, indicating that the model is highly accurate.

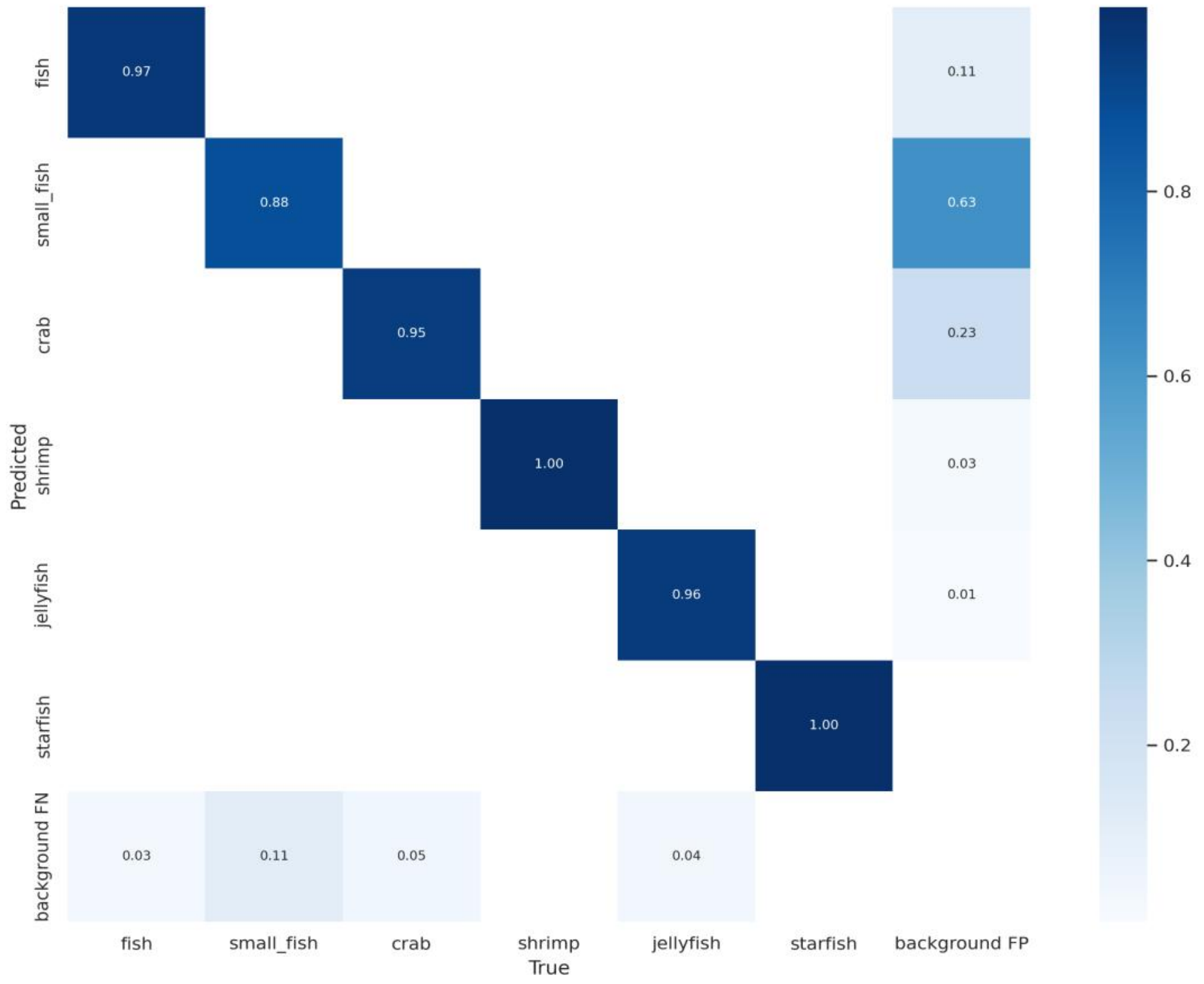

**Figure 17.** The confusion matrix of the YOLOv7-AC model (the Brackish dataset).

The variation curves of loss values with regard to the Brackish dataset were shown in Figure 18. As can be seen in Figure 18, the loss value steadily decreased and stabilized as the number of iterations increased.

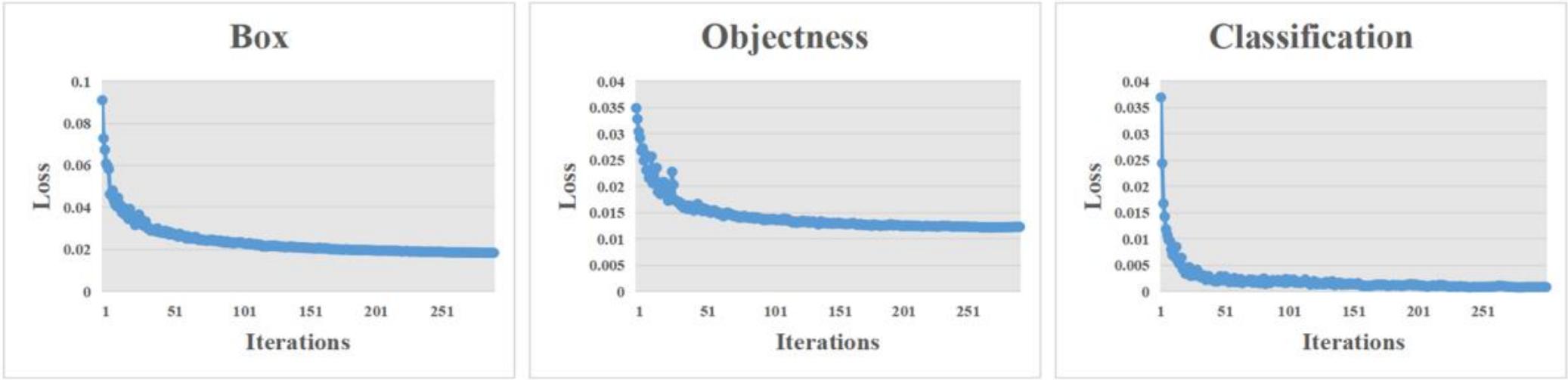

**Figure 18.** Variation curves of loss values on the Brackish dataset.

To further demonstrate the superiority of the proposed YOLOv7-AC model, a performance comparison was performed with popular target detection models on the Brackish dataset, where the correspondingly experimental results are shown in Table 6. As depicted in Table 6, the performance of the proposed YOLOv7-AC model was found to be superior to that of other popular target detection models, with mAP 1.1% higher than YOLOv7, and a respective improvement of 1.6%, 0.6%, and 7.5% compared to YOLOv6, YOLOv5s, and SSD. These experimental results demonstrate the clear advantages of the proposed method in the recognition of underwater targets.

**Table 6.** Performance comparison of target detection model on the Brackish dataset.

| Method | Precision | Recall | mAP@0.5 | mAP@0.95 |
|---|---|---|---|---|
| EfficientDet-d0[47] | 95.5% | 87.6% | 93.5% | 60.5% |
| SSD[21] | 91.2% | 79.9% | 89.9% | 53.7% |
| RetinaNet-50[48] | 86.9% | 76.4% | 85.2% | 49.1% |
| Detr[49] | 97.8% | 94.7% | 97.0% | 72.1% |
| YOLOv5s[27] | 96.1% | 94.3% | 96.8% | 70.2% |
| YOLOv6[28] | 95.6% | 92.6% | 95.8% | 65.8% |
| YOLOv7[29] | 96.3% | 93.7% | 96.3% | 73.2% |
| YOLOv7-AC | **98.2%** | **95.2%** | **97.4%** | **73.7%** |

4.4.3. Ablation experiments of the Brackish dataset

Accordingly, the ablation experiments were conducted to observe the effectiveness of different improvements on the model performance on the Brackish dataset, where the experimental results were shown in Table 7. As can be observed from Table 7, the application of each individual improvement leads to a relatively modest increase in performance. The integration of the ResNet-ACmix module and the AC-E-ELAN module resulted in a 0.2% and 0.4% increase in the mAP, respectively. Furthermore, the incorporation of GAM and the utilization of K-means++ clustering anchor boxes resulted in a 0.4% and 0.1% increase in the mAP, respectively, as seen from the pre-experiments.

**Table 7.** Ablation comparison of model performance improvement on the Brackish dataset.

| Model | ResNet-ACmix | AC-E-ELAN | GAM | K-means++ | AP（fish） | AP（small_fish） | AP（crab） | AP（shrimp） | AP（jellyfish） | AP（starfish） | mAP |
|---|---|---|---|---|---|---|---|---|---|---|---|
| YOLO v7 | × | × | × | × | 96.0% | 84.9% | 98.5% | 99.3% | 94.6% | 99.5% | 96.3% |
| | √ | × | × | × | 98.0% | 90.5% | 97.8% | 99.4% | 92.5% | 99.6% | 96.5% |
| | √ | √ | × | × | 98.0% | 90.2% | 99.2% | 99.2% | 95.5% | 99.5% | 96.9% |
| | √ | √ | √ | × | 98.2% | 91.5% | 99.3% | 99.1% | 95.5% | 99.5% | 97.3% |
| | √ | √ | √ | √ | 98.2% | 92.4% | 99.3% | 99.5% | 95.6% | 99.5% | 97.4% |

*4.5. The speed comparison of YOLOv7-AC and other models*

The performance of the proposed YOLOv7-AC model in terms of speed was evaluated by comparing its FPS metric with the popular target detection models applied to the URPC and Brackish datasets for training and testing. The experimental results, as shown in Table 8, indicate that the YOLOv5s model achieved the highest FPS score on both datasets, with YOLOv7-AC ranking second, slightly higher than YOLOv7, and significantly faster than the other models. These results demonstrate that the proposed YOLOv7-AC model not only offers improved accuracy, but also exhibits a noteworthy level of efficiency.

**Table 8.** Target detection model FPS comparison of the URPC dataset and the Brackish dataset.

| Method | The URPC Dataset | The Brackish Dataset |
|---|---|---|
| EfficientDet-d0[47] | 36 | 51 |
| SSD[21] | 51 | 77 |
| RetinaNet-50[48] | 23 | 46 |
| Detr[49] | 33 | 50 |
| YOLOv5s[27] | **77** | **101** |
| YOLOv6[28] | 64 | 86 |
| YOLOv7[29] | 73 | 90 |
| YOLOv7-AC | **74** | **92** |

## 5. Discussion

The challenges associated with detecting targets in harsh underwater scenes can be attributed to the issues of color distortion and low visibility caused by medium scattering and absorption in underwater optical images. To address these challenges, this study proposes the innovative use of the ACmix module, the design of the ResNet-ACmix module and the AC-E-ELAN module based on ACmix, along with the incorporation of the GAM, to enhance the extraction of informative features. The results of the experiments demonstrate the efficacy of the proposed YOLOv7-AC model in harsh underwater scenarios, as indicated by its improved performance compared to the traditional YOLOv7. This is demonstrated through a comparison of the detection results of YOLOv7 and YOLOv7-AC on the URPC dataset and the Brackish dataset, as illustrated in Figure 19. As demonstrated by this figure, the proposed YOLOv7-AC model outperforms the YOLOv7 model in terms of error detection and omission detection. Not only is a higher number of targets accurately detected, but the prediction boxes are also more precise.

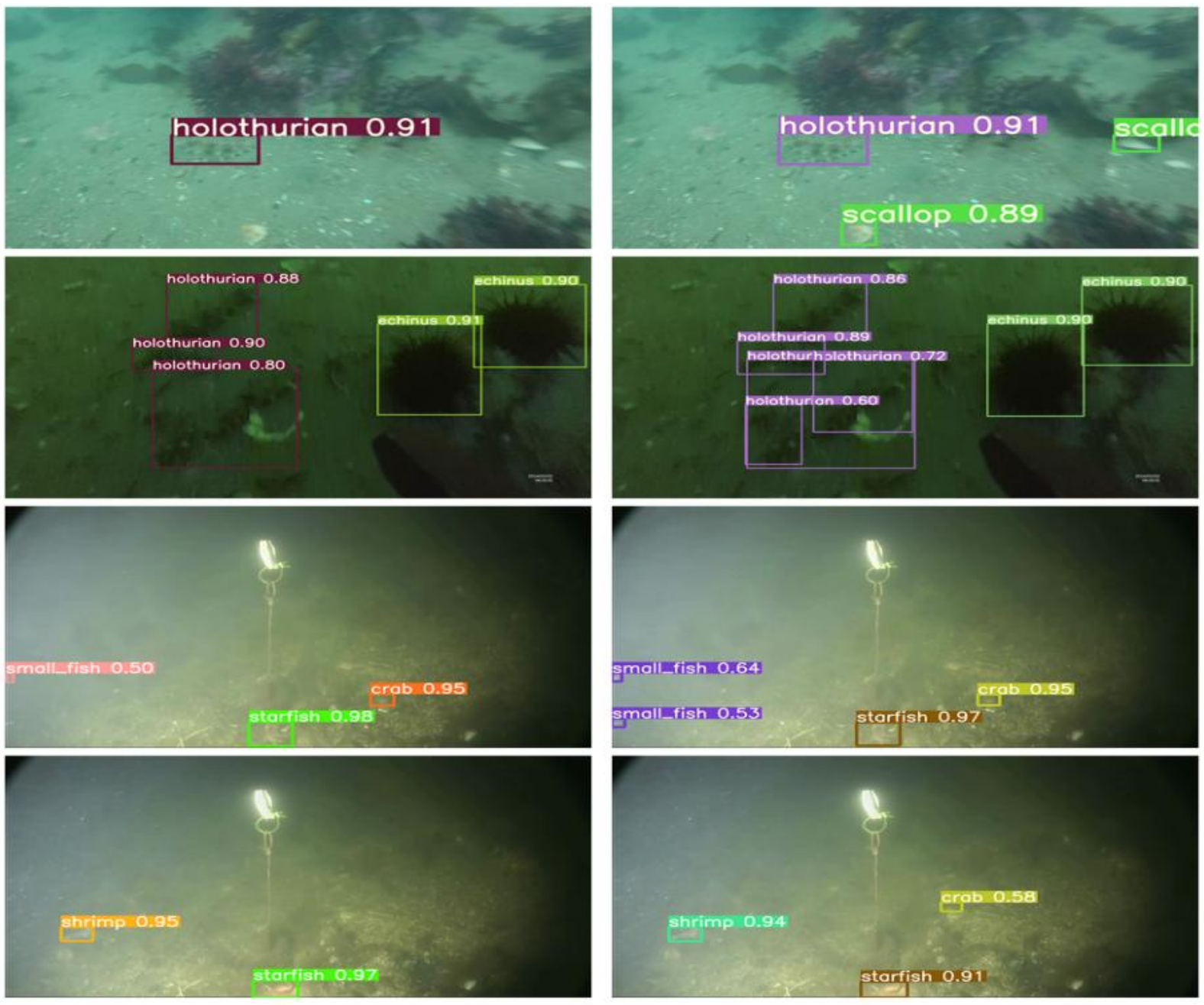

**Figure 19.** Detection results of YOLOv7 (**left**) and YOLOv7-AC (**right**) in harsh underwater scenes.

However, despite the improved performance, the YOLOv7-AC model still exhibits instances of false detection and missing detection in highly complex underwater environments. This can be observed in the examples presented in Figure 20.

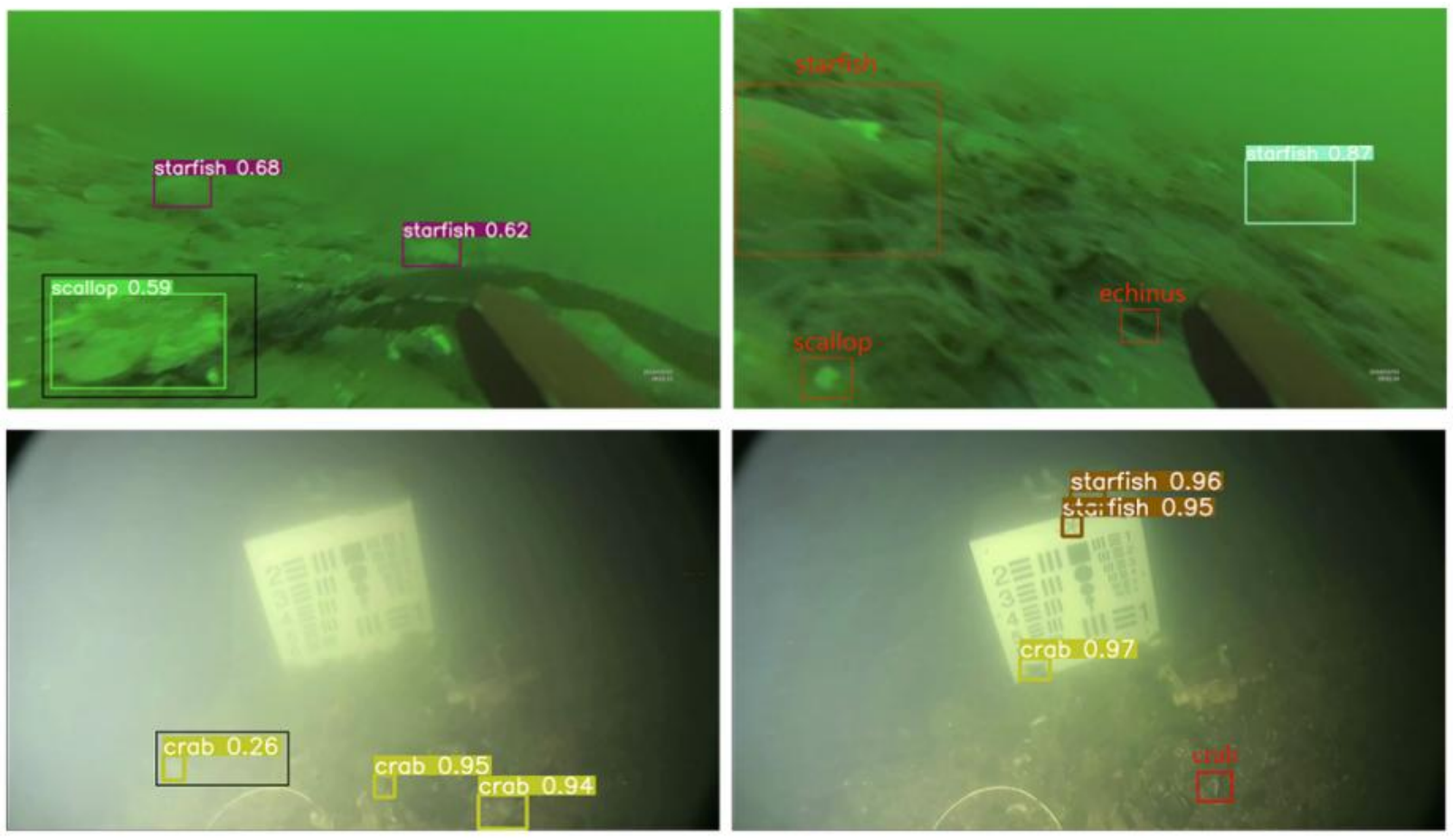

(a) (b)

**Figure 20.** (**a**) Error detection of YOLOv7-AC in highly complex underwater environments (**left:** marked in black boxes); (**b**) omission detection of YOLOv7-AC in highly complex underwater environments (**right**: marked in red boxes).

## 6. Conclusion

In this study, an improved YOLOv7-based network, referred to as YOLOv7-AC, is presented for the purpose of detecting targets in complex underwater environments. To achieve this, the AC-E-ELAN module is designed to emphasize target features, while the incorporation of jump connections and a 1x1 convolutional structure within the ACmixBlock improves computational speed and memory utilization. The ResNet-ACmix module is further developed to extract deep features that are more effectively trained by the network. Furthermore, the use of GAM and K-means++ enhances the overall performance of the detection. Experiments were conducted using the URPC and Brackish datasets, and the results were compared to those obtained using popular target detection algorithms and the proposed YOLOv7-AC model. The results indicate that the proposed YOLOv7-AC model surpasses the state-of-the-art target detection models in terms of its robustness and performance in complex underwater environments.

However, it must be noted that the availability of high-quality underwater datasets and images remains a major challenge in the development of target detection in underwater environments. Hence, the future research efforts will aim at collecting a large and diverse set of underwater datasets and employing image enhancement techniques to improve the overall quality of underwater images, which are crucial for the detection of underwater targets.